\pdfoutput=1

\documentclass[11pt]{article}

\usepackage[final]{acl}

\usepackage{times}
\usepackage{latexsym}
\usepackage{cleveref}
\usepackage[T1]{fontenc}

\usepackage[utf8]{inputenc}

\usepackage{microtype}

\usepackage{inconsolata}

\usepackage{booktabs}
\usepackage{hyperref}
\usepackage{graphicx}
\usepackage{xspace}
\usepackage{cleveref}
\usepackage{linguex}
\usepackage{tcolorbox}
\usepackage{multirow}
\usepackage{makecell}

\newtcolorbox[auto counter]{examplebox}[2][]{%
label type=Example,
title=Example~\thetcbcounter: #2,#1}

\title{Evaluating the Robustness of Adverse Drug Event Classification Models Using Templates}

\author{Dorothea MacPhail$^1$, David Harbecke$^1$, Lisa Raithel$^{1,2,3}$, Sebastian Möller$^{1,2}$ \\
  $^1$German Research Center for Artificial Intelligence (DFKI), Berlin \\
  $^2$Quality \& Usability Lab, Technische Universität Berlin \\
  $^3$BIFOLD – Berlin Institute for the Foundations of Learning and Data\\
  $^1$\texttt{\{firstname\}.\{lastname\}@dfki.de} \\
}

\begin{document}
\maketitle
\begin{abstract}
An adverse drug effect (ADE) is any harmful event resulting from medical drug treatment.
Despite their importance, ADEs are often under-reported in official channels.
Some research has therefore turned to detecting discussions of ADEs in social media.
Impressive results have been achieved in various attempts to detect ADEs.
In a high-stakes domain such as medicine, however, an in-depth evaluation of a model's abilities is crucial.
We address the issue of thorough performance evaluation in English-language ADE detection with hand-crafted templates for four capabilities: Temporal order, negation, sentiment, and beneficial effect.
We find that models with similar performance on held-out test sets have varying results on these capabilities.
\end{abstract}

\section{Introduction}\label{sec:Intro}

When a trained model is applied to real-world data, it may be confronted with phenomena that are under-represented or non-existent in the training data \cite{Belinkov-Bisk, MORADI-samwald}.
This raises the question of how to evaluate a model's performance and generalization abilities. 
Reporting summary statistics and held-out test set performance is a common practice in model evaluation.
While this can provide an indication of the model’s performance and ability to generalize, there are some issues with this practice.
Firstly, held-out test sets often arise from the same distribution as the training data and will, therefore, exhibit the same patterns and biases to a high degree.
Real-world data, however, may have different feature distribution or exhibit noise.
Held-out testing, therefore, often provides an unsatisfactory estimation of a model’s performance and generalization abilities \cite{Belinkov-Bisk, mccoy-etal-2019-right-for-wrong-reasons, ribeiro-etal-2018-SEARS}. 

Secondly, a high model score does not necessarily reveal what the model has learned during training.
Research has shown that a model may not learn relevant patterns but instead base its decisions on shallow heuristics or proxies \cite{mccoy-etal-2019-right-for-wrong-reasons}. 
Benchmark challenges have attempted to address this issue by testing models on a wide range of aspects of language \cite{superglue}.
However, not all aspects can be tested in a benchmark, and the benchmark itself may exhibit unintended biases \cite{kiela-etal-2021-dynabench}, so the question of what a model has learned remains.

Inspired by the behavioral testing suite CheckList \citep{checklist-acl20}, we propose the use of template-based test cases to test different capabilities of adverse drug effect (ADE) classification models.
ADEs are any harmful consequence to a patient due to medical drug intake.
Due to the potential detrimental outcomes of ADEs, the detection of ADEs is an important goal in health-related NLP and has been a subject of research for a considerable time. 
We test models in understanding of temporal order, positive sentiment, beneficial effects and negation (see Table \ref{tab:testcase-examples}).

\begin{table*}[ht!]
\small
\centering
\begin{tabular}{@{}llll@{}}
\textbf{Test Name} & \textbf{Label} & \textbf{Test Description} & \textbf{Example Test Case} \\ \midrule
\multirowcell{2}[0pt][l]{Temporal Order\\ standard} & no ADE & ADE occurs before drug intake & Before taking \underline{cymbalta}, I experienced \underline{Insomnia}. \\ \cmidrule{2-4}
 & ADE & ADE occurs after drug intake & Before having \underline{acid reflux}, I was put on \underline{zoloft}. \\ \midrule
\multirowcell{2}[0pt][l]{Temporal Order \\ single time entity} & no ADE & \begin{tabular}[c]{@{}l@{}}ADE occurs before drug intake \\ expressed by a time entity\end{tabular} & \begin{tabular}[c]{@{}l@{}}I was experiencing \underline{bad pain in my right arm} \\ for \underline{2 weeks}, now I started being medicated \\ with \underline{Effexor XR}.\end{tabular} \\ \cmidrule{2-4}
 & ADE & \begin{tabular}[c]{@{}l@{}}ADE occurs after drug intake \\ expressed by a time entity\end{tabular} & \begin{tabular}[c]{@{}l@{}}\underline{3 months} ago I started being treated with \\ \underline{zoloft}, now I started encountering \\ \underline{excellerated heart rate}.\end{tabular} \\ \midrule
\multirowcell{2}[0pt][l]{Temporal Order \\ double time entities} & no ADE & \begin{tabular}[c]{@{}l@{}}ADE occurs before drug intake \\ expressed by two related \\ time entities\end{tabular} & \begin{tabular}[c]{@{}l@{}}\underline{3 weeks} ago I started suffering from \underline{bad pain} \\ \underline{in my right arm}, I have been taking \underline{effexor} \\ for \underline{2 days}.\end{tabular} \\ \cmidrule{2-4}
 & ADE & \begin{tabular}[c]{@{}l@{}}ADE occurs after drug intake \\ expressed by two related \\ time entities\end{tabular} & \begin{tabular}[c]{@{}l@{}}I was enduring \underline{Insomnia} for \underline{6 weeks}, \\ \underline{8 weeks} ago I started taking \underline{cymbalta}.\end{tabular} \\ \midrule
Positive Sentiment & ADE & \begin{tabular}[c]{@{}l@{}}ADE occurrence is reported \\ with positive sentiment\end{tabular} & \begin{tabular}[c]{@{}l@{}}I'm taking \underline{cymbalta} and experiencing \\ \underline{cravings for sweets}. Still, I am happy \\ my symptoms have reduced.\end{tabular} \\ \midrule
Beneficial Effect & no ADE & \begin{tabular}[c]{@{}l@{}}Secondary effect of a drug \\ that is beneficial to the patient\end{tabular} & \begin{tabular}[c]{@{}l@{}}I'm taking \underline{Effexor XR} and experiencing \\ weight loss. I'm happy because I was trying \\ to lose weight anyway.\end{tabular} \\ \cmidrule{2-4}
 & ADE & \begin{tabular}[c]{@{}l@{}}Secondary effect of a drug that \\ is an ADE as it is not beneficial\end{tabular} & \begin{tabular}[c]{@{}l@{}}For me, weight loss is a side-effect of \\ \underline{effexor}. It's a problem because I am \\ already underweight.\end{tabular} \\ \midrule
Negation & no ADE & ADE is negated & \begin{tabular}[c]{@{}l@{}}I am taking \underline{zoloft} without suffering from \\ \underline{acid reflux}.\end{tabular} \\ \cmidrule{2-4}
 & ADE & \begin{tabular}[c]{@{}l@{}}Statement contains negation, \\ ADE is not negated\end{tabular} & \begin{tabular}[c]{@{}l@{}}That's not true, I took \underline{zoloft} and \\ encountered \underline{Insomnia}.\end{tabular} \\ \bottomrule
\end{tabular}
\caption{Overview of all four capabilities tested with example test cases.
The temporal order capability has three variations.
All test cases have an assigned label, either \texttt{ADE} or \texttt{no\;ADE}.
Filled-in entities are underlined in the example test cases.
All test cases are hand-crafted.}
\label{tab:testcase-examples}
\end{table*}

In high-stakes domains such as medicine, an in-depth evaluation of a model’s abilities is crucial.
Related work (\Cref{sec:RelatedWork}), however, suggests that shortcomings towards selected linguistic phenomena and reliance on proxies for model decisions may exist in models in the biomedical domain.

In this work\footnote{The templates and code can be found at \url{https://github.com/dfki-nlp/ade_templates}}, two transformer-based models for the detection of ADEs in user reports on social media were fine-tuned and tested by conventional held-out testing as well as additional template-based tests.
The results of held-out testing and the template-based tests were compared in order to better understand (i) the models' shortcomings and (ii) the potential gaps in knowledge that can occur when a model's abilities are only evaluated via test set performance.
We find that models underperform on some capabilities and show differences in some capabilities \textit{despite highly similar $F_1$-scores on the held-out test set}.
We therefore provide the following contributions:
\begin{itemize}
    \item A curated test bench of 99 templates with 1505 variations to investigate the robustness of ADE classification models across four capabilities.
    \item A comparison of two popular transformer-based models on long-tail linguistic phenomena in the classification of ADEs.
\end{itemize}

\section{Related Work}\label{sec:RelatedWork}

Studies on the detection of ADEs in user-generated texts have been conducted since approximately \citeyear{leaman_towards_2010}, when \citeauthor{leaman_towards_2010} published the first English dataset within this domain. 
The usual downstream tasks are those common in information extraction: Document classification, to find relevant documents containing mentions of adverse effects; named entity recognition, to identify medication and disease-related mentions; and relation classification, to establish associations between the entity mentions.
Approaches for all of these tasks range from rule- and lexicon-based systems \citep{leaman_towards_2010, nikfarjam_pattern_2011} to traditional machine learning pipelines \citep{gurulingappa_development_2012, ginn_mining_2014, segura-bedmar_detecting_2014} and, recently, deep neural networks \citep{huynh_adverse_2016}, specifically transformer-based setups \citep{weissenbacher_overview_2019, miftahutdinov_kfu_2020, gusev_bert_2020, magge_deepademiner_2021}. 

However, even advanced models struggle with the supposedly simple task of classifying a document into either ``contains an ADE'' (henceforth \texttt{ADE}) or ``does not contain an ADE'' (\texttt{no\;ADE}), a standard binary classification that is still necessary to find relevant documents for further information extraction. 
This is often due to a strong class imbalance (in most cases, the documents containing ADEs are in the minority), the usual noise in social media data, ambiguities in health-related statements of patients, and general weaknesses of language models in coping with certain linguistic phenomena not only with respect to ADEs. 

For example, \citet{scaboro-etal-2021-nade} have studied the extraction of ADEs from tweets using BERT, SpanBERT \citep{joshi2020spanbert}, and PubMedBERT \citep{gu2021domain}.
They tested all three models' ability to handle negation and detect shortcomings in all three models.
\citet{MORADI-samwald} investigated the robustness of four transformer models specialized in the biomedical and clinical domain over a variety of tasks such as sentence classification, inference, and question answering.
The models' robustness is tested by adding minor meaning-preserving changes to the input with the goal of fooling the model.
Their findings highlight the vulnerability of state-of-the-art transformer-based models to adversarial input.

Finally, there is CheckList \cite{checklist-acl20}, a model-agnostic framework aimed at testing a trained model's behavior and gaining an in-depth understanding of its potential shortcomings.
CheckList guides the creation of test cases based on natural language \textit{capabilities}, which are used as new inputs to the trained model and subsequently evaluated.
The idea is to determine which capabilities (e.g., negation handling, robustness) are necessary for the task the model is intended to perform.
\citet{checklist-acl20} identify three possible test types which can be used for testing the capabilities:
the Minimum Functionality Test (MFT), which targets a specific behavior similar to a unit test; the Invariance Test (INV), where the model’s robustness to irrelevant perturbations is tested; and the Directional Expectation Test (DIR), which consists of adding perturbations that are expected to lead to a specific outcome.
\citet{checklist-acl20} observe that the CheckList-based evaluation approach could not only uncover bugs in previously tested models but also that CheckList can make the search for bugs more systematic.
Recently, updates to CheckList, AdaTest \cite{ribeiro_2022_adatest} and AdaTest++ \cite{rastogi_2023_adatestplusplus}, were proposed which assist the user in finding bugs by suggesting topics and test cases in a semi-automated process.
While these are valuable additions, we decided to use the template-based approach for this project because we had pre-selected capabilities that we wanted to test with full control over the template design.

CheckList applications include the evaluation of general capabilities of models \cite{xie-etal-2021-regression} as well as evaluating models in specialized tasks such as offensive speech detection \cite{bhatt-etal-2021-case, manerba-tonelli-2021-fine} and automatic text simplification \cite{cumbicus-pinenda-checklist-ats}.
For the specialized tasks, the authors use CheckList to guide their testing approach by defining new capabilities specific to the task at hand.

In the biomedical and clinical domain, \citet{Ahsan2021MIMICSBDHAD} use CheckList to test four linguistic capabilities (negation, temporal order, misspellings, and attributions) on their transformer-based model with a dataset of clinical discharge notes.
One of their findings is that the model struggles to correctly distinguish between past and present mentions of substance use in the discharge notes.
The detection of ADEs, however, is not part of the research. 

The exposure of potential weaknesses in transformer-based models in the biomedical domain motivates an in-depth analysis of models used for ADE detection.
To our knowledge, a systematic template-based approach to test model capabilities has not yet been applied to ADE detection.

\section{Methods}\label{sec:Methods}

We use templates to test a selection of linguistic capabilities of binary ADE classification models.
To this end, we first manually create templates (see \Cref{sec:template_creation}) and then sets of test cases, by using entities to fill placeholders in the templates (see \Cref{sec:entities}).
We then evaluate two fine-tuned classification models on these test cases and compare their predictions with each other and with the models' performance on the held-out test set.

\begin{center}
    \small
    \begin{examplebox}[label type=example, label={ex:temporder_std}]{\small Template for Temporal Order (\texttt{ADE})}
    {\small I started taking \texttt{\{drug\}} before I experienced \texttt{\{ade\}}.}
    \end{examplebox}
\end{center}

We test four capabilities: \textit{Temporal Order}, \textit{Positive Sentiment}, \textit{Beneficial Effect}, and \textit{Negation} (see \Cref{sec:capabilities}).
99 base templates are created with 1505 variations (for details see Table \ref{tab:count_templates} in  \Cref{app:templates}).
Each template is also assigned a label (\texttt{ADE}/\texttt{no\;ADE}) in accordance with published guidelines for the annotation of ADEs (see \Cref{sec:finetune_data}).
The template in \Cref{ex:temporder_std} provides a test case for the capability \textit{Temporal Order} and has a positive label (\texttt{ADE}).
Filled-in template examples for every capability we test are listed in Table \ref{tab:testcase-examples}.
The filled-in templates (test cases) serve as the input to the fine-tuned model for inference.
In the following, we present more details about the template creation and the investigated capabilities. 

\subsection{Template Creation}\label{sec:template_creation}
\begin{figure}[h]
    \centering
    \hspace*{-0.25cm}
    \includegraphics[scale=0.5]{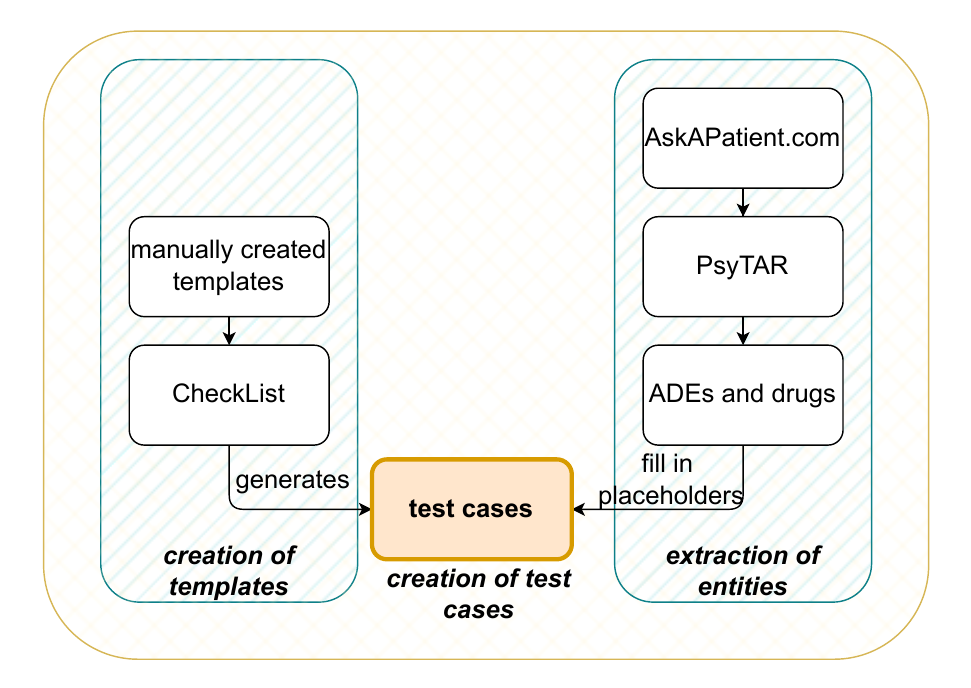}
    \caption{The process for creating test cases.}
    \label{fig:testcase_creation}
\end{figure}

Template-based evaluation is most effective with a large number of test cases that cover a diverse range of potential inputs.
These test cases are based on templates, which include placeholders.
For every placeholder, there is a list of potential entity fill-ins as in Example \ref{ex:temporder_std}, \texttt{\{drug\}} and \texttt{\{ade\}}, which could be filled with, e.g., \textit{Effexor} and \textit{nausea}.
The abstraction of test cases to templates allows to systematically capture important linguistic scenarios while creating a large number of different test cases. 
The process is visualized in \Cref{fig:testcase_creation}.

In the interest of linguistic diversity, variations of base templates were introduced for all capabilities except \textit{Beneficial Effect}. 
For \textit{Temporal Order} and \textit{Negation} templates, the vocabulary of the base template was modified to increase diversity.
\textit{Positive Sentiment} templates underwent syntax variations by exchanging or removing the conjunction between the two phrases.

The templates have a mean token count of 10.6 and 13.4 for the \texttt{no\;ADE} and \texttt{ADE} class respectively\footnote{Tokens were split at whitespace.}.
After filling in the entities for the placeholders, the average test case length in the experiments is 14.7 for the \texttt{no\;ADE} class and 16.6 for the \texttt{ADE} class.

\subsection{Capabilities}\label{sec:capabilities}
The choice of capabilities for this work is inspired by considerations on abilities a robust ADE classification model should possess and shortcomings of biomedical models as reported in \Cref{sec:RelatedWork}.
We based the phrasing of the templates on linguistic properties of social media posts: First-person usage, mostly single short sentences, and colloquial language.
Contractions were used occasionally.
However, usernames, misspellings, and non-standard grammar and punctuation were not applied in the templates as they manifest a separate capability.
All templates created can be viewed as templates for a CheckList Minimum Functionality Test \cite{checklist-acl20}.

To verify the existence of the described phenomena in the dataset, we randomly sampled 1,000 documents and let two annotators check each tweet for the occurrence of these phenomena.
The annotations showed that eight of the sampled tweets contain expressions of temporal order, one positive sentiment, one beneficial effect, and one negation. 
This sample showed that, as expected, the phenomena are rather rare but still exist in the long tail of the data distribution.
Nevertheless, an expert would expect a good classification model to have these capabilities.

\paragraph{Temporal Order}
The templates for testing \textit{Temporal Order} adapt the temporal structure test of \citet{checklist-acl20} and investigate the model's ability to correctly process information on past, present, and future as expressed in text. 
In the context of ADE detection, it is important for the model to ``understand'' temporal order since an effect cannot be an ADE if it occurred before the drug intake. 
According to the annotation guidelines based on which the data we use for fine-tuning was annotated, an effect occurring after a drug intake was labeled as ADE if the patient draws a connection between the effect and drug intake. 
Therefore, the templates assume an ADE when a harmful effect occurs after the drug intake. 

\paragraph{Positive Sentiment}

ADEs are often reported using negative sentiment \cite{alhuzali-ADE-sentiment}. 
If many ADE reports contain negative sentiment, an ADE detection model might perform well by using negative sentiment as a proxy. 
Nevertheless, a report might also be expressed favorably. 
This could be the case when a patient experiences relief from the original symptoms alongside a mild ADE. 
Therefore, an ADE detection model should recognize ADEs even when expressed in a positive framing so as not to miss out on less severe ADEs. 

\paragraph{Beneficial Effects}

The third capability is the correct distinction between ADEs and beneficial effects.
The latter are secondary effects of a drug that are not related to the reason for using the medication and which have, nevertheless, a positive outcome for the patient. 
Note that an effect may be regarded as positive or negative depending on the patient, their general health, and the context. 
Weight loss, for instance, may be considered a negative secondary drug effect or a beneficial effect depending on the patient.
The tests in this work assume that a positive secondary effect is a beneficial effect, not an ADE. 
The \textit{Beneficial Effect} test that expects a negative class label (\texttt{no\;ADE}) expresses the occurrence of a beneficial effect. 
The positive class (\texttt{ADE}) test consists of test cases that express an ADE that could be classified as a beneficial effect, but the context states that the user views the effect as negative. 

\paragraph{Negation}
\textit{Negation} templates test the model's ability to process negation in text.
Negation detection is a general challenge in NLP and a common phenomenon in language \cite{hossain-etal-2022-negation, truong-etal-2022-negation-benchmark}.
Thus, it is also an important capability for ADE detection.
The \textit{Negation} test that expects a negative class label (\texttt{no\;ADE}) contains a negated ADE.
The positive class (\texttt{ADE}) test cases include an ADE mention as well as a negation without negating the ADE.

\subsection{Entity Placeholders}\label{sec:entities}

All templates have entity placeholders for a drug name. 
Templates for \textit{Temporal Order}, \textit{Positive Sentiment}, and \textit{Negation} also have a placeholder for an ADE entity. 
Templates for \textit{Beneficial Effect} contain an effect that may be considered an ADE or a beneficial effect depending on the context. 
A list of the effects used in the \textit{Beneficial Effects} tests is provided in \Cref{sec:App-BenEff-list}. 
Template variations of the \textit{Temporal Order} capability that use time entities have placeholders for time expressions. 
The placeholders are filled with the respective time expressions from a self-created list of entities. 

\section{Experiments}\label{sec:Experiments}

We frame ADE detection as a binary classification task.
We first describe the experiments on the custom dataset and then the experiments on our template-based test cases.

\subsection{Fine-Tuning Experiments}
The following describes data, training and evaluation on the custom dataset.

\subsubsection{Data}\label{sec:finetune_data}
\begin{figure}[h]
    \centering
    \includegraphics[scale=0.37]{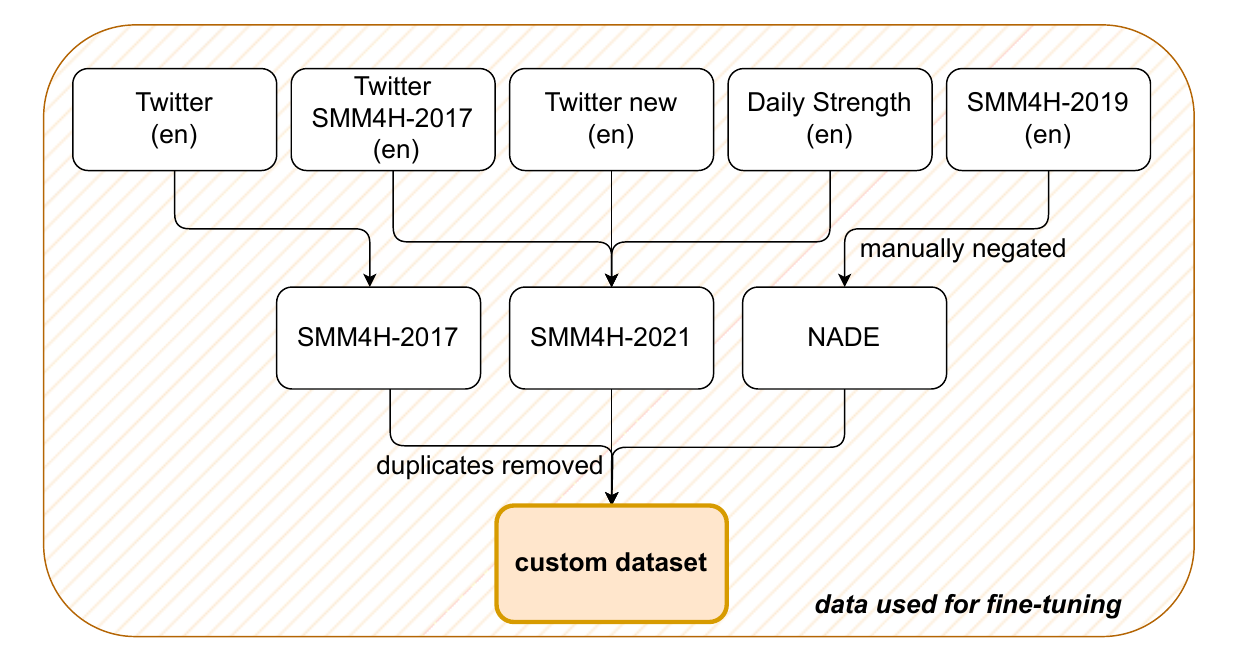}
    \caption{The different data sources for creating the custom dataset for fine-tuning the models.}
    \label{fig:custom_dataset}
\end{figure}

The custom dataset for our experiments consists of three social media corpora:
The SMM4H-2021 Shared Task 1a training data \citep{smm4h-2021-proceedings} (61\% of the custom dataset), the SMM4H-2017 Shared Task dataset \citep{smm4h-17-task} (38\%), and artificially negated tweets from the NADE dataset \citep{scaboro-etal-2021-nade} (1\%), resulting in 28,468 tweets. 
The data flow and their origin are shown in \Cref{fig:custom_dataset}.
Dataset statistics are covered in \Cref{tab:data-merged}.
In the user-reported texts, each sample either describes an ADE (\texttt{ADE}) or does not contain an ADE mention (\texttt{no\;ADE}).

\begin{table}
\small
\centering
\begin{tabular}{@{}lrr@{}}
\toprule
\textbf{Dataset} & \textbf{\#Tweets}  & \textbf{ADE Ratio (\%)} \\ \midrule
SMM4H'21 Task 1a                                                             & 17,426 & 7.39           \\
SMM4H'17 Task 1                                                                    & 14,880 & 8.72           \\
NADE  & 246   & 0.00              \\ \midrule
\textbf{Merged Dataset}                                                     & \textbf{28,468} & \textbf{8.75}           \\ \bottomrule
\end{tabular}
\caption{
The number of tweets per dataset and the respective ADE ratio (number of positive samples) of the merged dataset and its three components. 
4084 duplicates were removed after merging.
}
\label{tab:data-merged}
\end{table}

The SMM4H-2021 Shared Task 1a training data \cite{smm4h-2021-proceedings} itself consists of posts from Twitter and DailyStrength\footnote{\href{https://www.dailystrength.org/}{www.dailystrength.org}} collected using a list of 81 drugs widespread on the US market \citep{nikfarjam2015pharmacovigilance}. 
The data was annotated by two expert annotators. The annotators did not include beneficial effects 
in the ADE definition. 
It further includes some data previously used in the SMM4H-2017 Shared Task \cite{smm4h-17-task}. 

The SMM4H-2017 Shared Task data was collected from Twitter using generic drug names with a total of 250 keywords and subsequently annotated by two annotators. 
Again, the annotators excluded beneficial effects from the ADE definition.  
Overlapping texts between the SMM4H-2021 data and the SMM4H-2017 data used for our merged custom dataset were removed. 

The last part of our custom dataset
are artificially negated tweets from the NADE dataset \cite{scaboro-etal-2021-nade}. 
This dataset consists of tweets originating from the SMM4H-2019 Shared Task \citep{weissenbacher_overview_2019} and manually negated by annotators. 
Each negated tweet contains a statement that negates the presence of an ADE.
The three components are shown again in \Cref{tab:data-merged}.

We use this merged version of multiple datasets to give the fine-tuning models the best chance to learn different capabilities from varied data.
The texts in the custom dataset are between 1 and 34 tokens long.\footnote{Tokens were split at white spaces.} 
Negative (\texttt{no\;ADE}) samples are slightly shorter on average (16.2 tokens) than positive (\texttt{ADE}) samples (18.4 tokens).
These are slightly longer than our test cases with an average length of 14.7 tokens and 16.6 tokens.
Data splits for training, validation, and testing were created with a 70-10-20 ratio and stratified sampling by class label. 

\subsubsection{Model Fine-Tuning}
For the task of ADE classification, we fine-tune BioRedditBERT \citep{BioRedditBERT-cometa} and XLM-RoBERTa \citep{conneau2020unsupervised} on the custom dataset described in \Cref{sec:finetune_data}.
BioRedditBERT is a BERT-base uncased model related to BioBERT \cite{BioBERT-paper}, a model pre-trained on the original BERT training corpus (English Wikipedia + BookCorpus) as well as on medical texts sourced from PubMed and PMC.
It was then further fine-tuned on a corpus of health-related Reddit posts.
XLM-RoBERTa is a popular multilingual model with no specific medical pre-training data.
We chose these models to gain insights on robustness of a language model with medical knowledge compared with an general domain language model that has no specific medical knowledge.

The inputs were sampled with replacement weighted by class ratio due to the class imbalance.  
This sampling strategy resulted in a better $F_1$-score on the validation dataset.

\subsubsection{Held-Out Test Set Evaluation}

We evaluate the fine-tuned models on the test set using precision, recall, and $F_1$-score for each class.
The main metric we focused on is $F_1$ of the positive class due to the large class imbalance.
This metric was also used for hyperparameter tuning on the validation set.
We compare per-class recall to the models' performances on each capability of the test cases.
The goal of this comparison is to determine whether the template-based evaluation approach contradicts the overall impression of the model performance measured by held-out test set performance. 

\subsection{Test Case Experiments}

We use all templates for each test and randomly select only one template variation per base template for the capabilities \textit{Temporal Order}, \textit{Positive Sentiment}, and \textit{Negation} to have a manageable number of test cases.
We created a total of 11,265 test cases, of which 4,620 test cases belong to the negative class (\texttt{no\;ADE}) and 6,645 belong to the positive class (\texttt{ADE}). 
\Cref{tab:count-testcases} shows the number of test cases run per test. 

\begin{table}
\centering
\small
\begin{tabular}{@{}llr@{}}
\toprule
\textbf{Test} & \textbf{Label} & \textbf{\#Test Cases} \\ \midrule
Temporal Order & no ADE & 1,050 \\
\hspace{3mm} standard & ADE & 900 \\ \midrule
Temporal Order & no ADE & 1,050 \\
\hspace{3mm} single time entity & ADE & 1,050 \\ \midrule
Temporal Order & no ADE & 1,575 \\
\hspace{3mm} double time entities & ADE & 1,575 \\ \midrule
Positive Sentiment & ADE & 2,700 \\ \midrule
\multirow{2}{*}{Beneficial Effect} & no ADE & 120 \\
 & ADE & 120 \\ \midrule
\multirow{2}{*}{Negation} & no ADE & 825 \\
 & ADE & 300 \\ \midrule
\textbf{Total} & & \textbf{11,265} \\ \bottomrule
\end{tabular}
\caption{Number of test cases run per test. 
We have at least 120 test cases for each capability, so that we can expect our results to be representative.}
\label{tab:count-testcases}
\end{table}

A random sample of 15 ADEs, 15 mild ADEs, 5 drug names, 7 single time entities, and 7 relational time entities was taken. 
A list of sampled ADEs, mild ADEs, and drug names can be viewed in \Cref{sec:App-ExperimentDetails}.

\subsubsection{Drug and ADE Template Fill-Ins}

We need expressions of ADEs and medical drugs to fill in the placeholders in the templates.
These are automatically extracted from the PsyTAR dataset \cite{zolnoori2019systematic} of patient reports on psychiatric medications. 
The dataset consists of 891 Ask-a-Patient\footnote{\href{https://www.askapatient.com/}{www.askapatient.com}} patient forum posts on the topic of four psychiatric medications: Zoloft, Lexapro, Cymbalta, and Effexor XR.
The corpus was annotated for ADE mentions by four annotators with a health-related background. 
A mention was considered an ADE ``if there is an explicit report of any sign/symptom that the patient explicitly associated them with the drug consumption'' \cite{zolnoori2019systematic}. 
All four drug names of PsyTAR were extracted as well as two spelling variations of ``Effexor XR'' 
and lowercase versions of all drug names.
Statistics on the occurrences of the drug names in the custom training dataset can be found in Table \ref{tab:drugnames-traindata} in Appendix \ref{sec:App-ExperimentDetails}.
Extracting ADEs and drug names from the same domain ensures a high likelihood of compatibility between ADEs and medications. 

The ADE entities extracted from PsyTAR are user-generated descriptions of ADEs that are often multi-word expressions and which use non-standardized terms. 
We did not correct grammar and spelling errors in the extracted ADEs. 

We created the templates in a way that most short noun phrases\footnote{The longest extracted ADE has a length of 7 tokens.} fit as ADE entities, therefore, short noun phrases were filtered from the ADE mentions in PsyTAR. 
A total of 1,227 unique ADEs were extracted, amounting to 36.50\% of unique ADE entities in PsyTAR.\footnote{More details on the extraction can be found in \Cref{app:ade_extraction}.}

For the \textit{Positive Sentiment} test, the extracted ADEs were manually filtered to collect 60 less severe ADEs. 
This was a necessary step to avoid creating unrealistic test cases such as \textit{``I always have severe pain in my hands when I'm on Cymbalta, but I am happy my symptoms have reduced''}. 

The time entities for the variations in \textit{Temporal Order} tests were not extracted but generated. 
Numbers between 1 and 25 inclusive were combined with a noun (either ``days'', ``weeks'', or ``months''). 
A random selection of these combinations was used as time entities.  

\begin{table}[tb]
\centering
\small
\begin{tabular}{@{}llllll@{}}
\toprule
\textbf{Model} & \textbf{Class}  & \textbf{P} & \textbf{R} & $\mathbf{F_1}$ \\ \midrule
\multirow{2}{*}{BioRedditBERT} & \texttt{ADE}    & 0.720 & 0.676 & \textbf{0.698}\\
                               & \texttt{no\;ADE} & 0.969 & 0.975 & 0.972 \\ \midrule
\multirow{2}{*}{XLM-RoBERTa}   & \texttt{ADE}    & 0.720 & 0.681 & \textbf{0.700} \\
                               & \texttt{no\;ADE} & 0.970 & 0.975 & 0.972 \\
\bottomrule 
\end{tabular}
\caption{The results of the baseline models in precision (P), recall (R), and $F_1$-score on the test split. 
Positive class $F_1$ is highlighted as the most popular metric.
All scores are very close which would indicate that we can expect similar task understanding of the models.}
\label{tab:baseline_results}
\end{table}

\begin{figure*}[htb]
\centering
\includegraphics[width=\columnwidth]{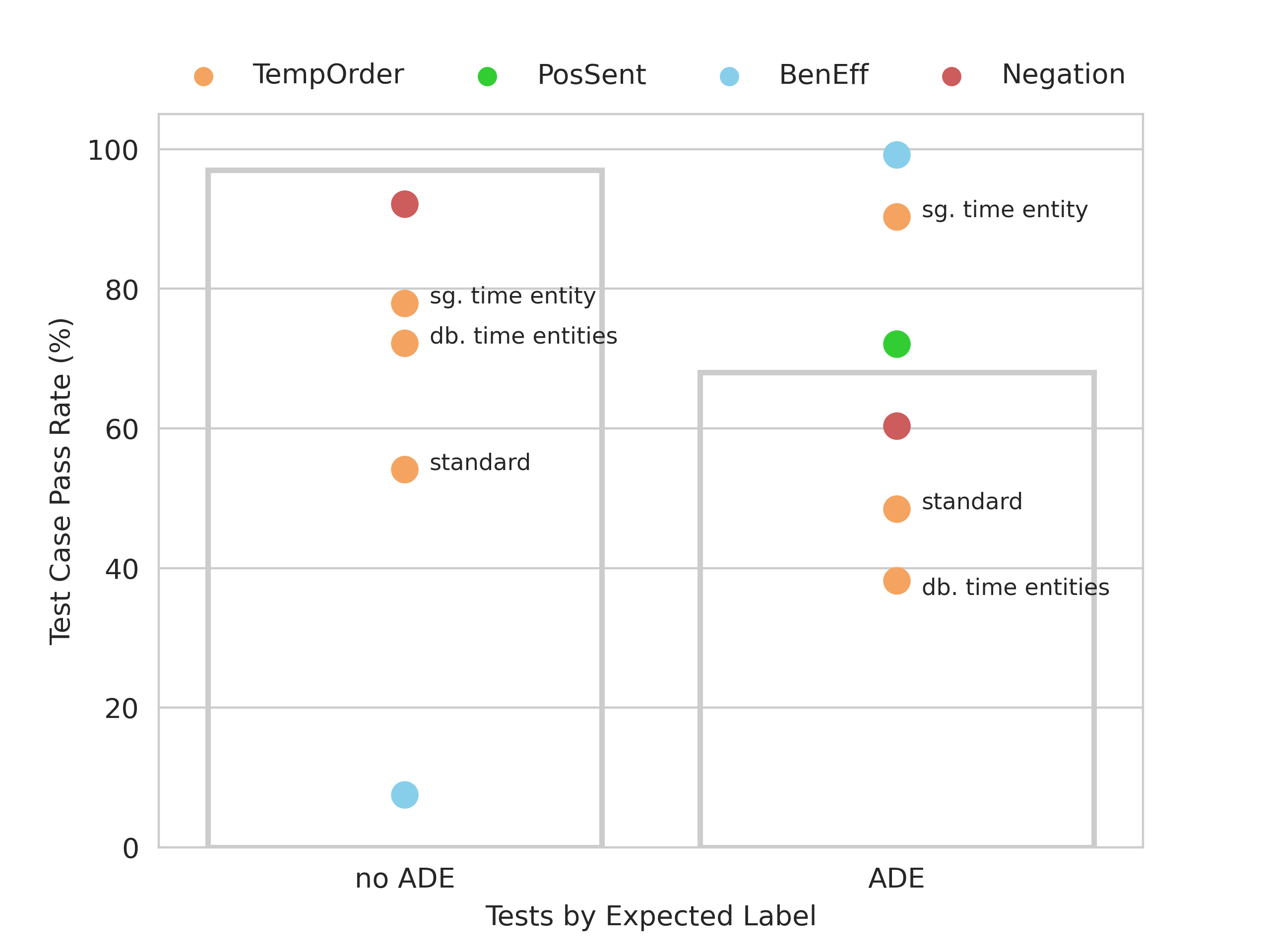}
\includegraphics[width=\columnwidth]{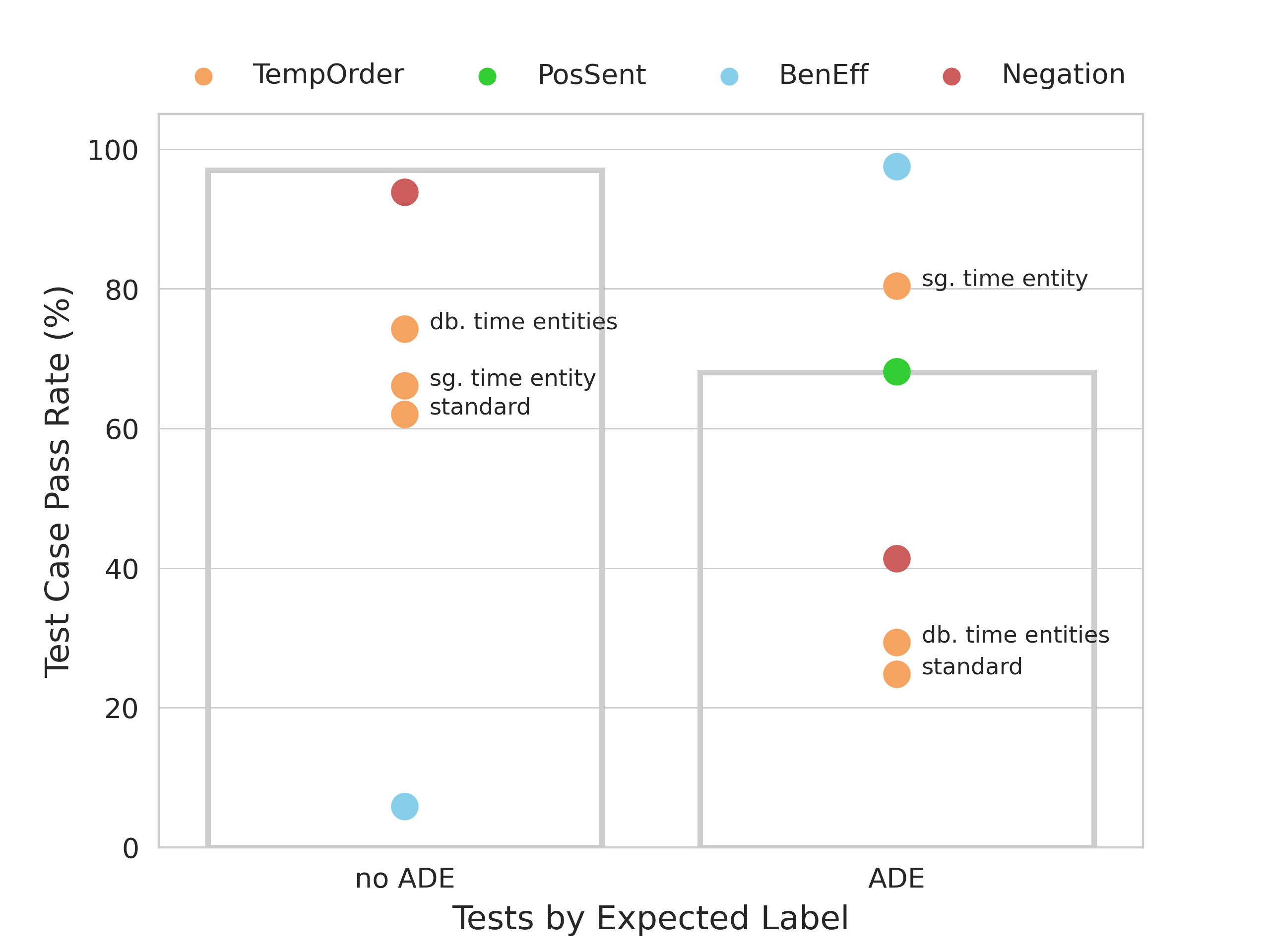}
\caption{Per-class performance of fine-tuned BioRedditBERT (left) and XLM-RoBERTa (right) on the test set (grey box baseline) and the capability-based test cases. The three distinct types of \textit{Temporal Order} tests refer to variety of \textit{Temporal Order} templates (\textit{standard}, \textit{single} and \textit{double time entity}) highlighted in Table \ref{tab:testcase-examples}. 
Most test cases are more difficult for the model to solve than the samples from the custom dataset.
The biggest difference between the models is the performance on the negation test cases with \texttt{ADE} label, where BioRedditBERT solves 20\% more test cases than XLM-RoBERTa.
Furthermore, both models have different performances for \textit{Temporal Order} test cases, especially standard cases with \texttt{ADE} label.}
\label{fig:results-dotplot}
\end{figure*}

\section{Results}\label{sec:Results}
The following presents the results of both the baselines and the template-based test cases. 

\subsection{Model Baseline}
The results of the baseline models can be found in \Cref{tab:baseline_results}.
All models were evaluated on the same test split of the fine-tuning corpus. 

The $F_1$-score of BioRedditBERT on the positive class (\texttt{ADE}) is 0.698, whereas XLM-RoBERTa achieves a score of 0.700, which indicates very similar general performance.
Due to the large class imbalance, the models reached a higher performance on the majority class (\texttt{no\;ADE}) with $F_1$-scores of 0.972.
The high overlap in data allows for comparison of this model's performance to the best performing models proposed in the latest SMM4H Shared Task on ADE classification \cite{smm4h-2022-proceedings}.

\subsection{Template-Based Test Results}

We compare model performance on the custom dataset to each template-based capability test performance separately.
Due to the variations in model performance over the two classes, we use per-class recall as a measurement of comparison between the model performance on the custom dataset and the template-based test cases as shown in \Cref{fig:results-dotplot}. 
For both models, all tests with \texttt{no\;ADE} labels fall short of the baseline performances. 
The highest level of performance is observed in the \textit{Negation} tests where BioRedditBERT and XLM-RoBERTa pass 92\% and 94\% of the test cases, respectively.
On the other hand, the \textit{Beneficial Effect} tests perform strikingly worse than the baselines with BioRedditBERT and XLM-RoBERTa passing only 7.5\% and 5.8\% of the test cases, respectively.
All three versions of the negative class \textit{Temporal Order} tests lie below the baselines but to a varying degree with a range of 54\%-78\% for BioRedditBERT and a range of 62\%-74\% for XLM-RoBERTa. 

For \texttt{ADE}, the models perform below the baseline (recall of 68\% for both models) on the \textit{standard Temporal Order} and \textit{double time entities Temporal Order} test (25\%-48\%), while the baseline is exceeded on the \textit{single time entity Temporal Order} test with 90\% for BioRedditBERT and 80\% for XLM-RoBERTa. 
Based on the varying model performance on different types of \textit{Temporal Order} tests for both the negative and the positive class, one can conclude that the model is not robust to changes in expression of temporal structure:
The use of single time entities affects the model performance positively compared to the use of prepositions (\textit{standard Temporal Order}) and double relational time entities.
Furthermore, BioRedditBERT (48\%) performs much better on \textit{standard Temporal Order} tests than XLM-RoBERTa (25\%).

Mild ADEs expressed in positive sentiment as in the \textit{Positive Sentiment} test do not pose a problem to the model. 
The performance on the \textit{Positive Sentiment} test cases (72\% for BioRedditBERT and 68\% for XLM-RoBERTa)
lies above the baseline of the positive class for both models. 
Also, the models' performance on the positive class negation test lies below the baseline, with BioRedditBERT (60\%) again performing much better than XLM-RoBERTa (41\%).

Unlike for the negative class test, almost all test cases in the \textit{Beneficial Effect} test on the positive class are correctly classified as \texttt{ADE}. 
The poor performance on the negative \textit{Beneficial Effect} test and the outstanding performance on the positive class \textit{Beneficial Effect} test leads to the conclusion that the model has not learned to distinguish between ADEs and beneficial effects.
Both models classify 96\% of Beneficial Effects test cases as \texttt{ADE}, even though half of the tests have a \texttt{no ADE} mention.
Possible explanations for this behavior are that the number of beneficial effect samples in the custom dataset is low and/or that the model does not take the context into account that distinguishes an ADE from a beneficial effect.  

\begin{figure}[t]
\centering
\includegraphics[width=\columnwidth]{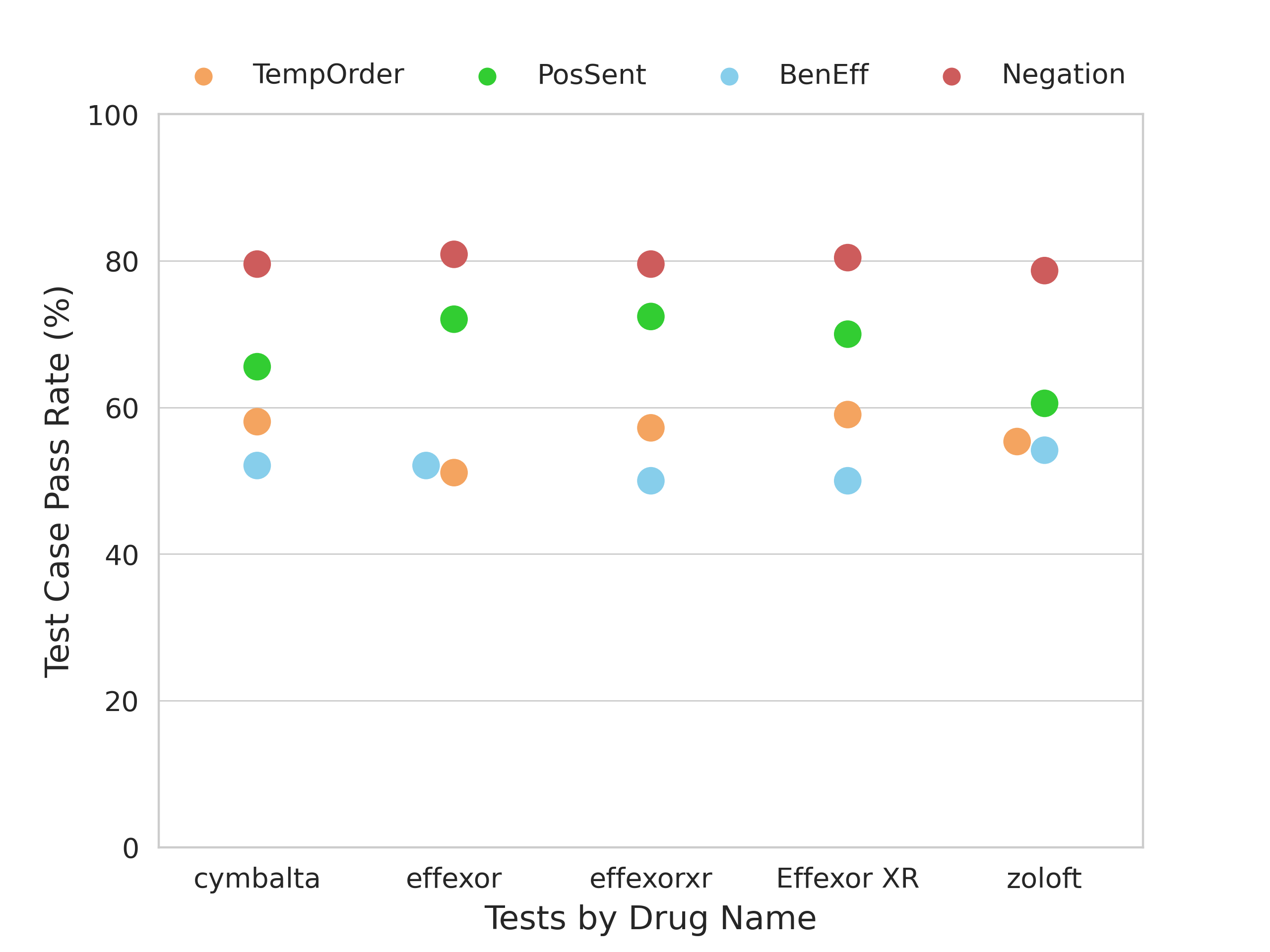}
\caption{Performance of XLM-Roberta on test cases by drug name and by capability. The number of test cases per capability and drug name is 1440 (Temporal Order), 540 (Positive Sentiment), 48 (Beneficial Effect), 225 (Negation).}
\label{fig:xlm_cl_bydrug_results}
\end{figure}

Each of the five selected drug name variants was used in every template allowing for an analysis of the impact of drug names in the test cases.
Performance variations on test cases with different drug names indicate reduced robustness of the model. 
We find slight variations in the model performance over different drug names as shown in figure \Cref{fig:xlm_cl_bydrug_results} for XLM-Roberta.
A potential explanation of these variations may be deviations in the occurrence of the respective drug names in the custom fine-tuning dataset, see Table \ref{tab:drugnames-traindata} in Appendix \ref{sec:App-ExperimentDetails}.

\section{Conclusion and Future Work}\label{sec:Conclusion}

In this work, we present a template-based approach for evaluating capabilities of models on the task of ADE detection in social media texts. 
Four capabilities, \textit{Temporal Order}, \textit{Positive Sentiment}, \textit{Beneficial Effect}, and \textit{Negation}, were identified and corresponding tests were created. 
Two high-performing models for the task of ADE detection were evaluated using the adapted approach.  

Results show that the models' performances vary across capabilities. 
While both models perform well on the \textit{Positive Sentiment} tests,
BioRedditBERT outperforms XLM-RoBERTa on \textit{Negation}.
The models are not able to distinguish between ADEs and beneficial effects and are not robust to changes in the expression of temporal structure in text. 
In summary, the template-based approach adapted to ADE classification has provided a better understanding of the shortcomings of high-performing models and can highlight previously undetected differences between models that perform almost identically on a held-out test set. We publish the templates to enable researchers to evaluate their own ADE classification models.

Further research may expand on this work by adding tests for more capabilities and evaluating other models using this approach. 
For example, in the phenomena annotation described in \Cref{sec:capabilities}, we found 1.6\% questions and 1.1\% speculative content in the tweets.
The linguistic variety of the templates could be improved by using a large language model to generate templates or test cases.
A different direction of research may focus on improving the model's faults detected during evaluation. 
One method of improvement is to include a subset of the test cases as new training data \citep{mccoy-etal-2019-right-for-wrong-reasons}.

\section{Limitations}\label{sec:Limitations}
While the approach of generating new inputs by templates undoubtedly has benefits, it also introduces some limitations.
For instance, the combination of all entity fill-ins with all templates can produce some unnatural phrases.
An example of this is the \textit{Temporal Order} template "After taking \{drug\}, I had \{ade\}.".
The ADE entity "weight gain" creates the unnatural sounding test case "After taking cymbalta, I had weight gain." instead of "After taking cymbalta, I gained weight."
The unnatural use of language may introduce a bias.
This should be kept in mind when using the templates.
However, not all entity fill-ins will introduce such a bias and the model's performance on the test cases cannot be fully attributed to the effect of unnatural language use.  

A second potential bias when using templates is that it may not be able to depict a large variety of language when only few templates were used.
An example of this are the templates for the positive class \textit{Beneficial Effect} test where each test case includes the word "problem".
A model could use this as a proxy for correctly classifying the test cases. 

Lastly, as described in \Cref{sec:Experiments}, not all features of social media tests were used when creating templates. No anonymized usernames, hashtags, non-standard punctuation, and colloquialisms other than contractions were applied in the templates. This may introduce a bias as there is a slight difference in language variety between the templates and the training data. A researcher should keep in mind that slight changes in the model performance may also be attributed to this shift in language variety. 

\section*{Acknowledgments}
We would like to thank the anonymous reviewers for their feedback on the paper. 
We further thank our annotators Alon Drobickij, Selin Yeginer, and Emiliano Valdes Menchaca.

This work was partially supported by the German Federal Ministry of Education and Research as part of the project TRAILS.
We are furthermore grateful for the support of the German Research Foundation (DFG) under the trilateral ANR-DFG-JST AI Research project KEEPHA (DFG-442445488), and the German Federal Ministry of Education and Research under the grant BIFOLD24B.
\bibliography{custom}

\appendix

\section{Templates}\label{app:templates}
The number of templates with linguistic variations for each capability can be seen in Table \ref{tab:count_templates}.
Example templates without filled-in entities are in Table \ref{tab:tests-templates}.

\begin{table}[htb]
\small
\centering
\begin{tabular}{@{}lrr@{}}
\toprule
\textbf{capability} & \multicolumn{1}{l}{\textbf{\#base templates}} & \multicolumn{1}{l}{\textbf{\#all templates}} \\ \midrule
TempOrder           & 36                                          & 816                                        \\
PosSent             & 36                                          & 504                                        \\
BenEff              & 12                                          & 48                                         \\
Negation            & 15                                          & 137                                        \\ \midrule
\textbf{Total}      & \textbf{99}                                 & \textbf{1505}                              \\ \bottomrule
\end{tabular}
\caption{Count of all created templates. Linguistic variation was used to create all templates from base templates.}
\label{tab:count_templates}
\end{table}

\begin{table*}[ht!]
\small
\centering
\begin{tabular}{@{}llll@{}}
\toprule
\textbf{Test Name} & \textbf{Label} & \textbf{Test Description} & \textbf{Example Template} \\ \toprule
\multirowcell{2}[0pt][l]{Temporal order\\ standard} & no ADE & ADE occurs before drug intake & Before taking \{drug\}, I experienced \{ade\}. \\ \cmidrule{2-4}
 & ADE & ADE occurs after drug intake & Before having \{ade\}, I was put on \{drug\}. \\ \midrule
\multirowcell{2}[0pt][l]{Temporal order \\ single time entity} & no ADE & \begin{tabular}[c]{@{}l@{}}ADE occurs before drug intake \\ expressed by a time entity\end{tabular} & \begin{tabular}[c]{@{}l@{}}I was experiencing \{ade\} for \{time\_entity\}, \\ now I started being medicated with \{drug\}.\end{tabular} \\ \cmidrule{2-4}
 & ADE & \begin{tabular}[c]{@{}l@{}}ADE occurs after drug intake \\ expressed by a time entity\end{tabular} & \begin{tabular}[c]{@{}l@{}}\{time\_entity\} ago I started being treated with\\ \{drug\}, now I started encountering \{ade\}.\end{tabular} \\ \midrule
\multirowcell{2}[0pt][l]{Temporal order \\ double time entities} & no ADE  & \begin{tabular}[c]{@{}l@{}}ADE occurs before drug intake \\ expressed by two related \\ time entities\end{tabular} & \begin{tabular}[c]{@{}l@{}}\{time\_entity\_large\} ago I started suffering from\\ \{ade\}, I have been taking \{drug\} \\ for \{time\_entity\_small\}.\end{tabular} \\ \cmidrule{2-4}
 & ADE & \begin{tabular}[c]{@{}l@{}}ADE occurs after drug intake \\ expressed by two related \\ time entities\end{tabular} & \begin{tabular}[c]{@{}l@{}}I was enduring \{ade\} for \{time\_entity\_small\}, \\ \{time\_entity\_large\} ago I started taking \{drug\}.\end{tabular} \\ \midrule
Positive Sentiment & ADE & \begin{tabular}[c]{@{}l@{}}ADE occurrence is reported \\ with positive sentiment\end{tabular} & \begin{tabular}[c]{@{}l@{}}I'm taking \{drug\} and experiencing \{ade\}. \\ Still, I am happy my symptoms have reduced.\end{tabular} \\ \midrule
Beneficial Effect & no ADE  & \begin{tabular}[c]{@{}l@{}}Secondary effect of a drug \\ that is beneficial to the patient\end{tabular} & \begin{tabular}[c]{@{}l@{}}I'm taking \{drug\} and experiencing weight loss. \\ I'm happy because I was trying to \\ lose weight anyway.\end{tabular} \\ \cmidrule{2-4}
 & ADE & \begin{tabular}[c]{@{}l@{}}Secondary effect of a drug that \\ is an ADE as it is not beneficial\end{tabular} & \begin{tabular}[c]{@{}l@{}}For me, weight loss is a side-effect of \{drug\}. \\ It's a problem because I am already underweight.\end{tabular} \\ \midrule
Negation & no ADE  & ADE is negated & I am taking \{drug\} without suffering from \{ade\}. \\ \cmidrule{2-4}
 & ADE & \begin{tabular}[c]{@{}l@{}}Statement contains negation, \\ ADE is not negated\end{tabular} & \begin{tabular}[c]{@{}l@{}}That's not true, I took \{drug\} \\ and encountered \{ade\}.\end{tabular} \\ \bottomrule
\end{tabular}
\caption{Overview of all CheckList tests conducted for this project with example templates.
Curly brackets in the example templates indicate entity placeholders.
}
\label{tab:tests-templates}
\end{table*}

\subsection{Extraction of ADEs from PsyTAR}\label{app:ade_extraction}

Sets of Parts of Speech combinations (tagsets) were created to define which sets of POS tags constitute a short noun phrase. 
An English POS tagger (spaCy) was then used to tag every token in the PsyTAR ADEs and filter out the chosen noun phrases. 
Examples of PsyTAR ADEs that were retrieved using this method are ``listlessness'', ``recurrence of ocular migraines'', and ``bad pain in my right arm''. 
The goal of this process was to retrieve as many and diverse ADE descriptions as possible, yet the tagsets are not extensive and not all relevant ADEs were retrieved. 
Reasons for not passing the tagset filters were not being a noun phrase (``gained 18 pound''), incorrect POS tag assigned tagger (``heartburn''), incorrect POS tags assigned due to typos or extra whitespace, long noun phrases (``stomach cramping the first couple of days''), and punctuation marks/symbols (``increase in alcohol abuse/dependence''). 

\subsection{List of Beneficial Effects}\label{sec:App-BenEff-list}
List of (potential) beneficial effects used for the \textit{Beneficial Effect} tests: weight loss/weight gain, sleepiness/decreased need for sleep, loss of appetite/increased appetite.

\section{Experiment Details}\label{sec:App-ExperimentDetails}
List of entities used as fill-ins for ADE, milder ADE for the \textit{Positive Sentiment} test, and drug names used in the experiments for this project.
\begin{itemize}
    \item drug names: zoloft, effexor, cymbalta, Effexor XR, effexorxr
    \item ADEs: Incredible sweet tooth, big appetite, many dreams, Difficulty Orgasming, excellerated heart rate, Insomnia, blackouts, bad pain in my right arm, a little more lethargy, VERY vivid dreams, stiff shoulders, EXTREME DRY MOUTH, Dialated pupils, increase in Libido, acid reflux
    \item milder ADEs: sugar craving, carbohydrate cravings, bouts of sleeplessness, gum pain, secretion under my toungue, weird dreams, stiff muscles, mild constipation, arm tingling, increased heat sensitivity, strange dreams, poorer concentration, cravings for sweets, hard time falling asleep, neck pain
\end{itemize}
The counts of the occurrences of the drug names can be found in Table \ref{tab:drugnames-traindata}.

\begin{table}[ht!]
\small
\centering
\begin{tabular}{@{}lrr@{}}
\toprule
\textbf{}  & \multicolumn{1}{l}{\textbf{exact matches}} & \multicolumn{1}{l}{\textbf{all matches}} \\ \midrule
cymbalta   & 451                                        & 742                                      \\
effexor    & 172                                        & 312                                      \\
effexorxr  & 0                                          & 0                                        \\
Effexor XR & 13                                         & 23                                       \\
zoloft     & 50                                         & 100                                      \\ \bottomrule
\end{tabular}
\caption{Occurrence of drug names in the fine-tuning training data. 
Exact matches are case-sensitive.
A sample can contain multiple drug name occurrences.
''effexorxr`` was used in the templates without appearing in the training data.}
\label{tab:drugnames-traindata}
\end{table}

\section{Model Details}\label{app:models}

\paragraph{BioRedditBERT \citep{BioRedditBERT-cometa}} is a BERT-base uncased model related to BioBERT \cite{BioBERT-paper}, a model pre-trained on the original BERT training corpus (English Wikipedia + BookCorpus) as well as on medical texts sourced from PubMed and PMC.
BioRedditBERT, in turn, was initialized from BioBERT and continued to pre-train on a corpus of health-related Reddit posts. The Reddit dataset contains 800.000 posts from 68 health-related subreddits collected between 2015 and 2018. The specific set of training data of BioRedditBERT was the pivotal argument for choosing this model for the task of ADE classification on the Twitter dataset.

\paragraph{XLM-RoBERTa \citep{conneau2020unsupervised}}
XLM-RoBERTa is a popular multilingual classification model without a focus on the biomedical domain. 

We conducted hyperparameter search for both models and tried batch sizes of $8$, $16$ and $32$ and learning rates of $3\cdot 10^{-6}$, $10^{-5}$ and $3\cdot 10^{-5}$.
Both models achieved the best performance on the development set at $16, 10^{-5}$ and trained with the AdamW \citep{loshchilov2017decoupled} optimizer.
No truncation of inputs was applied and the model was evaluated on the validation set after every epoch.
The inputs were sampled (batch sampling) with replacement weighted by class ratio due to the class imbalance (see \Cref{sec:finetune_data}). 

\section{Per-Template Performance}\label{app:pertemplate-perf}
\begin{figure*}[ht]
\centering
\includegraphics[scale=0.27]{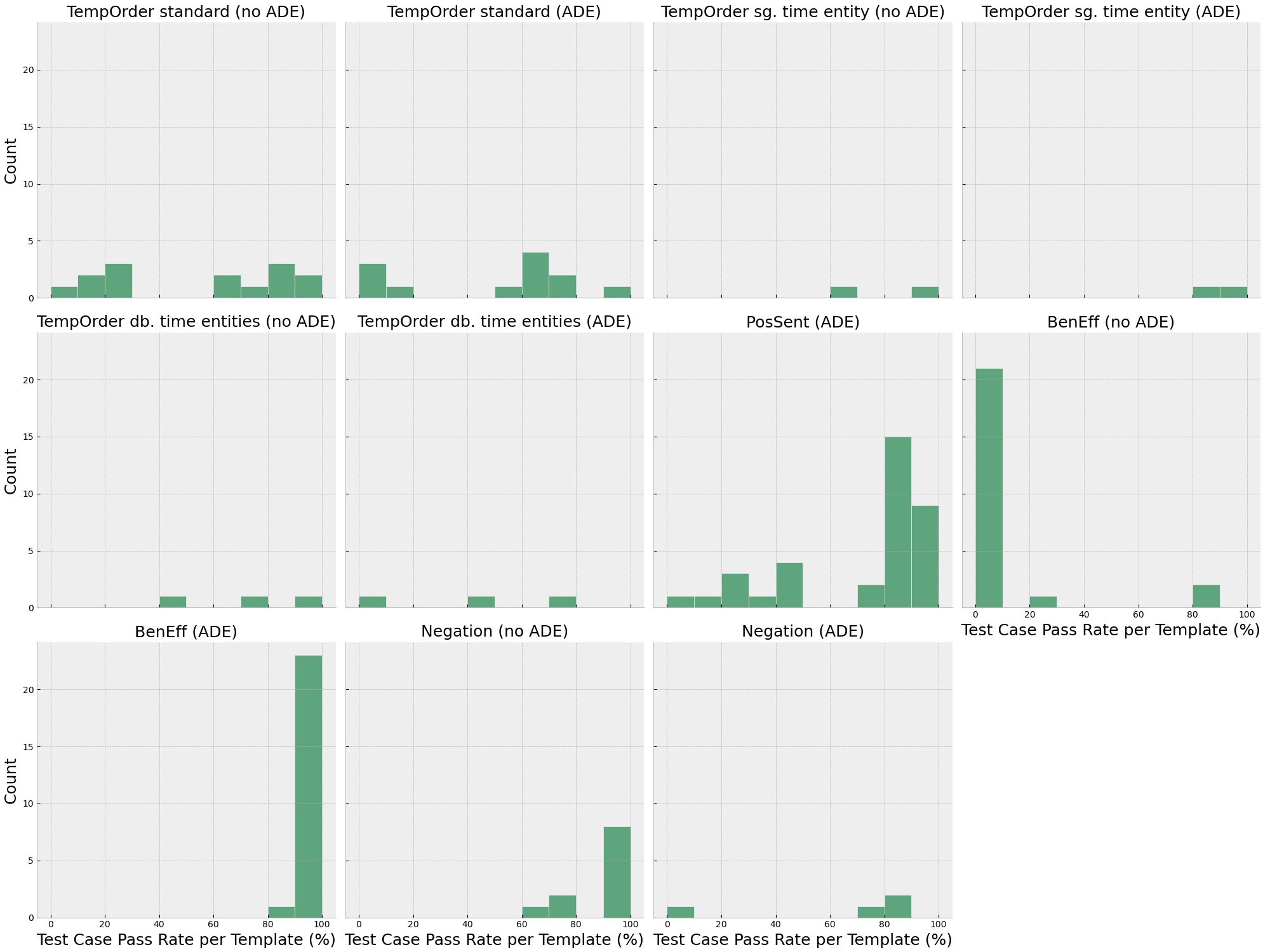}
\caption{Results of the CheckList tests on the fine-tuned BioRedditBERT by template. The ratio of correctly classified test cases per template is shown on the horizontal axis. Each plot is a histogram showing the count of templates that produced more or less successfully classified test cases.}
\label{fig:results_pertemplate_brb}
\end{figure*}

\begin{figure*}[ht]
\centering
\includegraphics[scale=0.27]{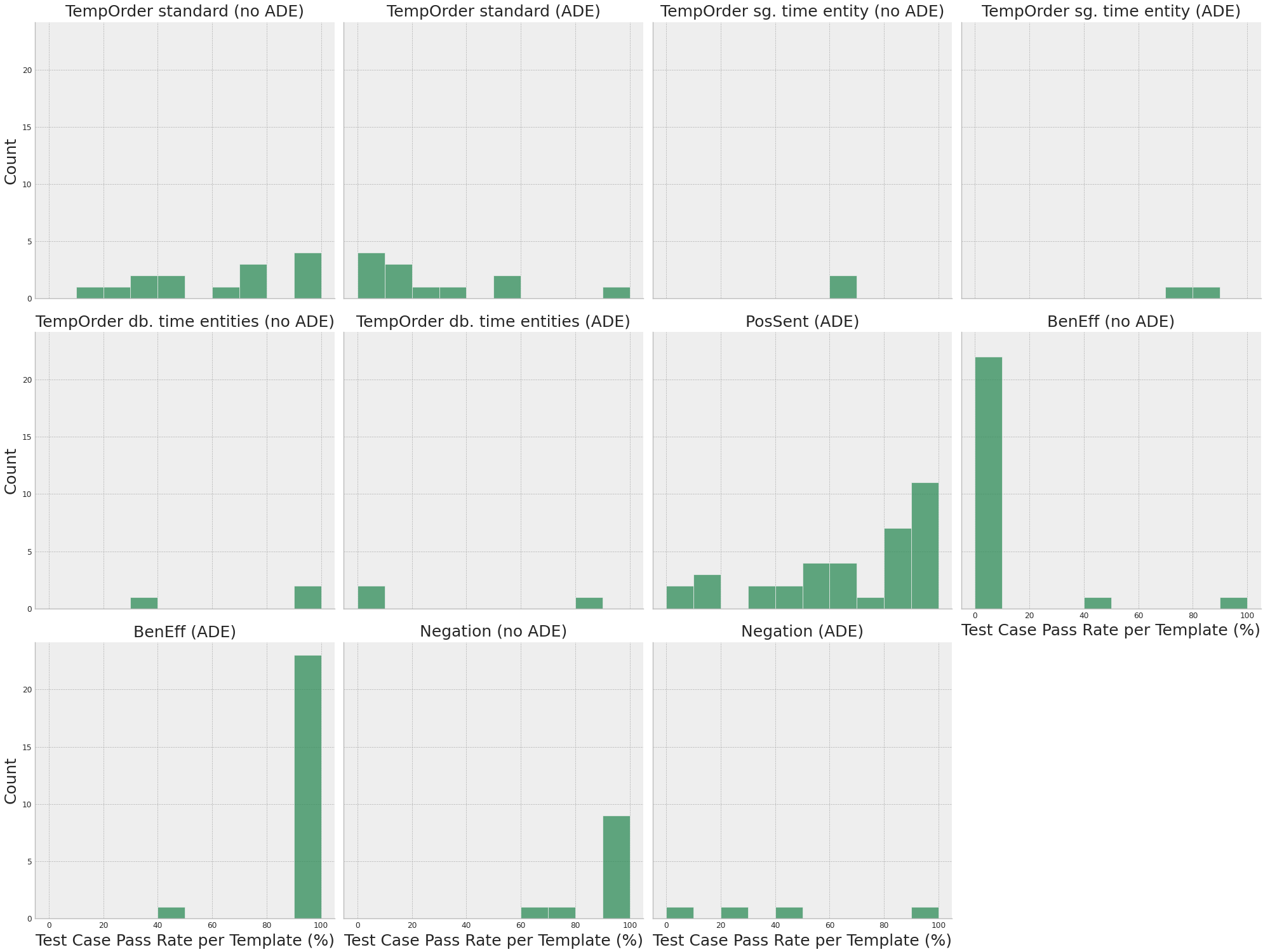}
\caption{Results of the CheckList tests on the fine-tuned XLM-RoBERTa by template. The ratio of correctly classified test cases per template is shown on the horizontal axis. Each plot is a histogram showing the count of templates that produced more or less successfully classified test cases.}
\label{fig:results_pertemplate_xlm}
\end{figure*}

The performance of the models on the template-based tests also varies within each test. 
For all tests except the \textit{Beneficial Effect} tests, the models' performance varies for each template, see Figures \ref{fig:results_pertemplate_brb} and \ref{fig:results_pertemplate_xlm}.
The dependence of the model performance on the template demonstrates that the wording of a template influences the models' ability to handle a capability. 
In turn, this stresses the importance of creating a wide range of variations in templates when using template-based evaluation. 

\end{document}